\documentclass{article}

\usepackage[numbers]{natbib}

\usepackage{hyperref}
\usepackage {arxiv}
\usepackage{cite}
\usepackage{floatrow}
\usepackage{algpseudocode} 
\usepackage{amsmath,amssymb,amsfonts}
\usepackage{caption}
\usepackage{algorithm}
\usepackage{textcomp}
\usepackage[table]{xcolor}
\usepackage{float}
\usepackage{booktabs}
\usepackage{enumitem}
\usepackage{setspace}
\usepackage[]{graphicx}
\usepackage{amsmath}
\usepackage{amssymb}
\usepackage{epsfig}
\usepackage{epstopdf}

\usepackage{subfigure}
\usepackage{caption}
\usepackage{comment}
\usepackage{multirow}

\title{\LARGE \bf
Deep Clustering of Tabular Data by Weighted Gaussian Distribution Learning}

\author{Shourav B. Rabbani, Ivan V. Medri, and Manar D. Samad\\
Department of Computer Science\\
 Tennessee State University\\
 Nashville, TN, USA\\
\texttt{msamad@tnstate.edu} \\
   }
   
\begin{document}



\maketitle

\begin{abstract}
Deep learning methods are primarily proposed for supervised learning of images or text with limited applications to clustering problems. In contrast, tabular data with heterogeneous features pose unique challenges in representation learning, where deep learning has yet to replace traditional machine learning. This paper addresses these challenges in developing one of the first deep clustering methods for tabular data: Gaussian Cluster Embedding in Autoencoder Latent Space (G-CEALS). G-CEALS is an unsupervised deep clustering framework for learning the parameters of multivariate Gaussian cluster distributions by iteratively updating individual cluster weights. The G-CEALS method presents average rank orderings of 2.9(1.7) and 2.8(1.7) based on clustering accuracy and adjusted Rand index (ARI) scores on sixteen tabular data sets, respectively, and outperforms nine state-of-the-art clustering methods. G-CEALS substantially improves clustering performance compared to traditional K-means and GMM, which are still de facto methods for clustering tabular data. Similar computationally efficient and high-performing deep clustering frameworks are imperative to reap the myriad benefits of deep learning on tabular data over traditional machine learning.
\end{abstract}


\section{Introduction}

Deep learning has replaced traditional machine learning in many data-intensive research and applications due to its ability to perform concurrent and efficient representation learning and classification. This concurrent learning approach outperforms traditional machine learning that requires \emph{handcrafted} features for classification~\citep{hand, ALAM2020}. However, representation learning via supervisory signals from ground truth labels may be prone to overfitting~\citep{prone_overfit} and adversarial attacks~\citep{prone_adversarial}. Moreover, human annotations for supervised representation learning and classification can be hard to obtain and unavailable in all data domains. Therefore, representation learning via deep unsupervised clustering may enable deep learning of vast unlabeled data samples available in practice. 

One of the ways to overcome the limitations of supervised representation learning is to generate pseudo labels via self-supervision, which does not require human-annotated supervisory signals~\citep{Boubekki2021, Caron2018}. A self-supervised autoencoder preserves input data information in a low-dimensional embedding for data reconstruction. However, the embedding resulting from a data reconstruction objective may not be optimal representations for downstream classification or clustering tasks~\citep{rabbani2024attention}. Therefore, deep learning methods have been jointly optimized with a clustering algorithm to obtain \emph {clustering friendly} representations~\citep{Xie2016, Guo2017, MoradiFard2020, Mrabah_neunet_2020, Yang2017}. The existing embedding clustering methods use traditional clustering algorithms (e.g., k-means) in joint optimization, assume t-distributed clusters, and benchmark on image data sets. While deep representation learning of images is well studied using convolutional neural networks (CNN), similar methods are not well developed for tabular data with heterogeneous feature space. The literature has strong evidence that traditional machine learning outperforms deep models in the supervised learning of tabular data~\citep{Kohler2019, Smith2020, Borisov2021, Kadra2021, Shwartz-Ziv2022}. However, deep learning methods have not been proposed for clustering tabular data based on a recent survey on deep clustering~\citep{zhou2022comprehensive}. This paper reviews the assumptions made in the embedding clustering literature to propose a novel deep clustering method for tabular data.

The remainder of this manuscript is organized as follows. Section 2 provides a review of the state-of-the-art literature on deep embedding clustering. Section 3 introduces tabular data and some theoretical contrasts between the embeddings for data visualization and clustering to lay the foundation for our proposed method.
Section 4 outlines the proposed deep clustering framework for effective cluster representation and assignments. Section 5 summarizes the tabular data sets and experiments for evaluating the proposed deep clustering method. Section 6 provides the experimental results and compares the clustering performances between the proposed and nine state-of-the-art clustering methods. Section 7 summarizes the findings with additional insights into the results. The paper concludes in Section 8.

\section{Related work}

A recent survey article reviews deep clustering methods for image, text, video, and graph data without any examples applied to tabular data sets~\citep{zhou2022comprehensive}. One of the earliest methods for embedding clustering, Deep Embedded Clustering (DEC)~\citep{Xie2016}, is inspired by the seminal work on t-distributed stochastic neighborhood embedding (t-SNE)~\citep{VanDerMaaten2008}. The DEC method first trains a deep autoencoder by minimizing the data reconstruction loss. The trained encoder part (excluding the decoder) is then fine-tuned by minimizing the Kullback-Leibler (KL) divergence between t-distributed clusters on the embedding (Q) and a target distribution (P).  The target distribution is a closed-form mathematical expression obtained by taking the derivative of the KL divergence loss with respect to P and equating it to zero. Therefore, the target distribution (P) is also a function of t-distributed Q in similar work. 
Later, the k-means clustering in the DEC approach is replaced by spectral clustering to improve the embedding quality in terms of clustering performance~\citep{Duan2019}. 

The DEC method is also enhanced by an improved DEC (IDEC) framework~\citep{Guo2017}. In IDEC, the autoencoder reconstruction and the KL divergence losses are jointly minimized to train a pre-trained deep autoencoder.  Similar strategies, including t-distributed clusters, k-means clustering, and KL divergence loss, are adopted in joint embedding and cluster learning (JECL) for multimodal representation learning of text-image data pairs~\citep{Yang2020}. The Deep Clustering via Joint Convolutional Autoencoder (DEPICT) approach learns image embedding via a de-noising autoencoder~\citep{Dizaji2017}. Unlike earlier methods, the DEPICT method proposes a clustering head with the softmax function to obtain soft cluster assignments without a distribution assumption. However, their method aims to achieve balanced clusters when cluster imbalance is a common problem with tabular data. They demonstrate that a cross-entropy loss can replace the KL divergence in minimizing the difference between P and Q distributions.

The embedding clustering literature commonly uses cluster assignments from k-means in a deep learning framework~\citep{Xie2016, Guo2017, Mrabah_neunet_2020, MoradiFard2020, Zhang2019, Yang2020, Dizaji2017}. The assumption of t-distributed cluster embedding made in the DEC method~\citep{Xie2016} continues to appear in subsequent studies~\citep{Ren2019, Enguehard2019, Guo2017, Duan2019, Yang2020, Wu2022, sadeghi2021idecf, zhou2022comprehensive}. A t-distribution is parameterized only by cluster centroid, whereas a multivariate Gaussian distribution can be used to learn both cluster centroid and covariance. Furthermore, the t-distribution assumption is originally made for neighborhood embedding for the t-SNE data visualization algorithm~\citep{VanDerMaaten2008}. We argue that the distribution assumption for data visualization may not optimally satisfy the requirements for clustering.

Furthermore, recent deep clustering methods optimized and improved on image data sets alone may not be optimal or even suitable for tabular data with heterogeneous feature space~\citep{van2020scan, li2021contrastive, sadeghi2021idecf, li2022_histopathological_clustering, sadeghi2022c3, zhao2023_semi_gmm, Yu2024_semi_clustering, sadeghi2024deep}. 
Some of these models use CNN-based large learning architectures for clustering large image data sets~\citep{van2020scan, li2021contrastive, sadeghi2022c3}. 
However, these large image-based CNN architectures are not suitable for learning tabular data sets with heterogeneous features. 
Some of these methods are sensitive to the values selected for multiple hyperparameter~\citep{sadeghi2022c3, sadeghi2024deep}. In several studies, class labels are leveraged to identify an early stopping point while pretraining the autoencoder, which may violate the unsupervised nature of clustering algorithms~\citep{sadeghi2021idecf, sadeghi2024deep}. A few methods also used class labels in a semi-supervised step to perform clustering~\citep{zhao2023_semi_gmm, Yu2024_semi_clustering}, which may go against the unsupervised nature of clustering algorithms. There is a need for deep clustering methods for tabular data addressing these methodological shortcomings.


\subsection{Contributions}

This paper proposes the first method for deep embedding clustering of tabular data to the best of our knowledge by addressing the shortcomings of state-of-the-art deep clustering methods. First, we replace the current assumption of t-distributed embedding with a mixture of multivariate Gaussian distributions for multivariate tabular data by providing a theoretical underpinning for this choice. Second, a new embedding clustering algorithm is proposed that can learn the cluster embedding and assignments without requiring the aid of a traditional clustering method. Third, multivariate cluster centroids and covariance matrices are updated as trainable parameters, and the cluster distributions are adjusted by individual cluster weights to learn imbalanced tabular data better. Fourth, a deep autoencoder directly learns cluster distributions using a dynamic distribution target instead of setting a mathematically closed-form target distribution or a KL divergence loss. 



\section {Theoretical background}

This section provides preliminaries on tabular data compared to commonly used benchmark image data sets. We draw multiple contrasts between the neighborhood embedding proposed for data visualization and the deep embedding required for clustering to underpin our proposed method.

\subsection {Preliminaries}

A tabular data set is represented in a matrix X $\in \Re^{n\times d}$ with $n$ i.i.d samples in rows. Each sample ($X_i$) is represented by a d-dimensional feature vector, $X_i \in \Re^{d} = \{x_1, x_2, …, x_d\}$, where i = $\{1, 2,…,n\}$. Compared to a homogeneous pixel distribution $P (I)$ of an image $I$,  tabular data contain multivariate distributions $P(x_1, x_2, …, x_d)$ of heterogeneous features in relatively much lower dimensions. One popular example of tabular data is Electronic Health Records (EHR), where individual patient follow-ups are characterized by heterogeneous clinical variables (e.g., heart rate, blood pressure, height)~\citep{kazijevs2023deep}. EHR data have been used to cluster patients of varying risk levels using one of the deep clustering methods (IDEC)~\citep{kowsar2023deep}. Tabular data are also used in missing value imputation literature, where cluster labels are used to facilitate the imputation of missing values in heterogeneous samples~\citep{samad2022missing}.

Table~\ref{imageVStable} shows several important contrasts between image and tabular data. One may argue that some high-dimensional sequential data, such as genomics or two-dimensional images converted to pixel vectors, can be structured in a data table. However, these vector representations still include regularity or homogeneity in patterns that do not pose the challenge of heterogeneity of tabular data. Therefore, tabular data with a heterogeneous feature space fail to take advantage of deep learning methods because image-like sequential or spatial regularities are absent. Furthermore, the current literature selectively uses data sets with high dimensionality and large sample sizes to demonstrate the effectiveness of deep learning methods~\citep{Arik2021Tabnet, Kossen2021Npt, Gorishniy2021Ftt}. In contrast, the most commonly available tabular data sets have limited sample sizes and dimensions (Table~\ref{imageVStable}), which are rarely considered in deep representation learning. Consequently, tabular data sets are identified as the last \emph {unconquered castle} for deep learning~\citep{Kadra2021}, where traditional machine learning methods still appear competitive against deep neural network architectures~\citep{Kadra2021, Borisov2021}.

\begin{table}[h]
\centering 
\caption{Contrasts between image and tabular data require different deep learning architectures for tabular data. Median sample size and data dimensionality are obtained from the 100 most downloaded tabular data sets from the UCI machine learning repository \citep{Dua_2019}.}
\label{imageVStable}
\scalebox{0.85}{
\begin{tabular}{lcc}
\toprule
Factors      & Image data & Tabular data               \\ \midrule
\multirow{2}{*}{Heterogeneity} & Homogeneous          & Heterogeneous                   \\
&pixels& variables\\
Spatial Regularity & Yes & No  \\
Sample size   & Large, $>$50,000         & Small, median$\sim{660}$        \\

Benchmark data sets     & MNIST, CIFAR         & None               \\
Data dimensionality     & High, $>$1000      &   Low, median 18        \\
\multirow{2}{*}{Best method}  & \multirow{2}{*}{Deep CNN}         & Traditional           \\
&&machine learning \\
\bottomrule
\end{tabular}
}
\end{table}

\subsection {Embedding for data visualization}

\begin{figure*}[t]
\centering
\vspace{-10pt}
\subfigure[t-SNE projected] { \includegraphics[width=0.40\textwidth] {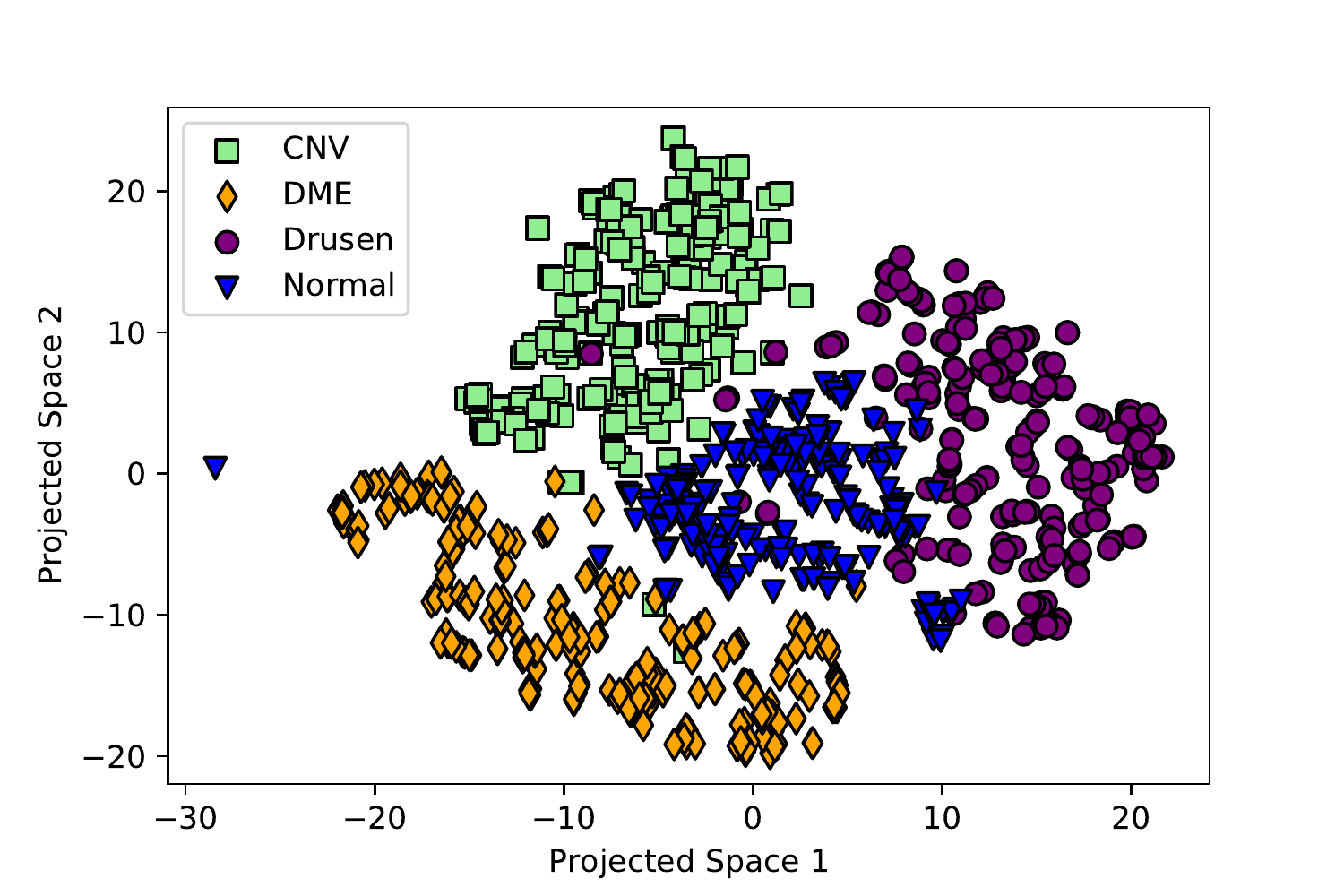}}
\subfigure[PCA projected] { \includegraphics[width=0.40\textwidth]  {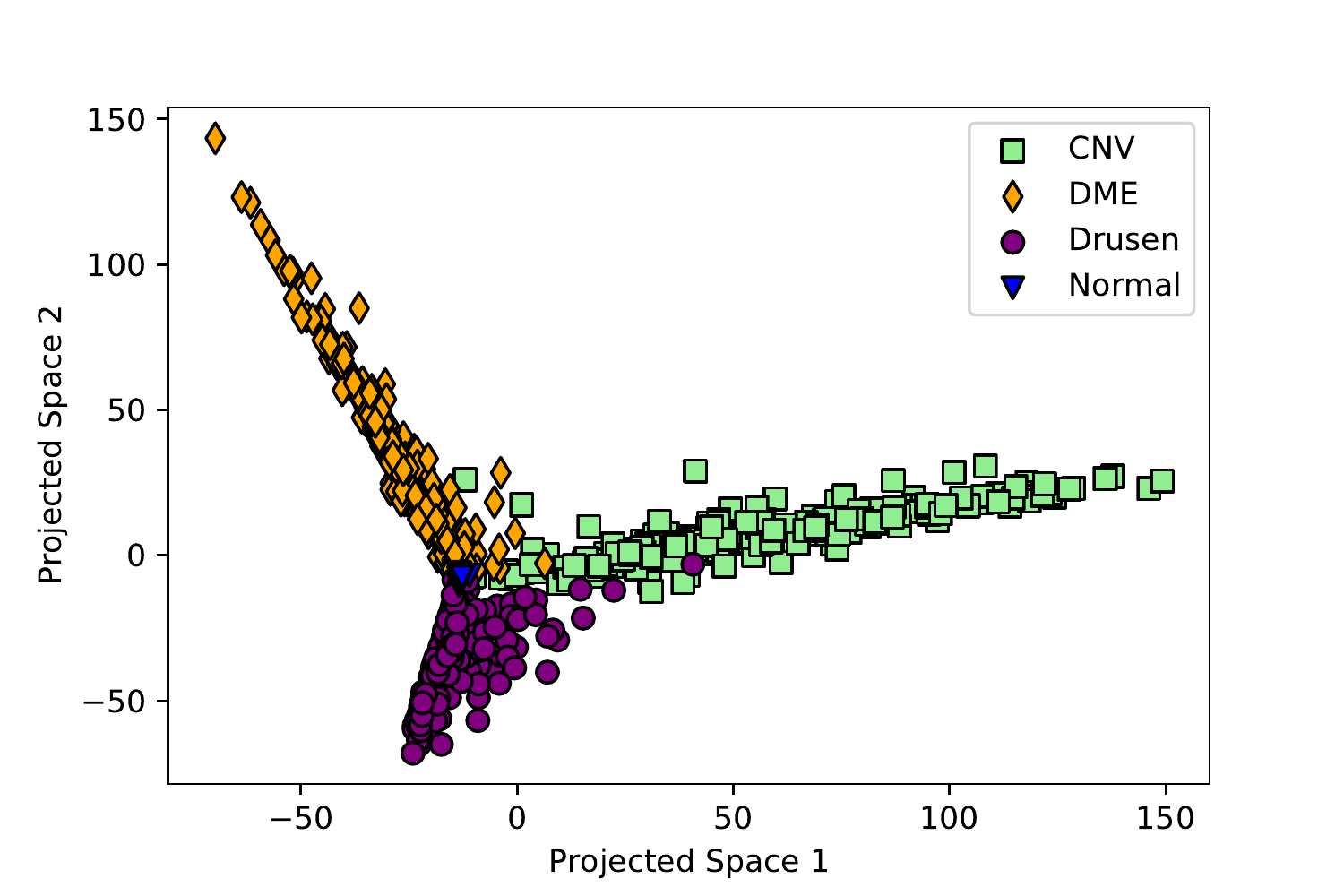}}
\vspace{-10pt}
\caption{Two-dimensional embeddings of high dimensional image features extracted from a deep convolutional neural network obtained from~\citep{Arefin2021}.} 
\label{t-SNE-PCA}
\end{figure*}

A neighborhood embedding is a low-dimensional map that preserves the similarity between data points ($x_i$ and $x_j$) observed in a higher dimension. Maaten and Hinton propose a Student's t-distribution to model the similarity between samples in neighborhood embedding ($z_i$, $z_j$) of high-dimensional data points ($x_i$ and $x_j$) for data visualization~\citep{VanDerMaaten2008}. First, the similarity between two sample points ($x_i$ and $x_j$) in the high dimension is modeled by a Gaussian distribution, $p_{ij}$ in Equation 1. Similar joint distribution can be defined for a pair of points in the low-dimensional embedding ($z_i$, $z_j$) as $q_{ij}$ in Equation 2.
\begin{eqnarray}
p_{ij} =  \frac{\exp ({-||x_i - x_j||^2/2 \sigma^2})}{\sum_{k\neq l}\exp ({-||x_k - x_l||^2/2 \sigma^2})},\\
q_{ij} =  \frac{\exp ({-||z_i - z_j||^2/2 \sigma^2}}{\sum_{k\neq l}\exp ({-||z_k - z_l||^2/2 \sigma^2})}
\end{eqnarray}
Notably, the low-dimensional embedding here is not obtained via a neural network as seen in deep embedding clustering. The divergence between the target ($p_{ij}$) and embedding ($q_{ij}$) distributions is measured using a KL divergence loss, which is minimized to optimize the final t-SNE embedding iteratively. 
\begin{equation}
    KL~(P || Q) = \sum_i \sum_j p_{ij} log \frac{p_{ij}}{q_{ij}}\\
\end{equation}
To facilitate high-dimensional data visualization in two dimensions (2D), the embedding distribution ($q_{ij}$) is modeled by a Student's t-distribution in Equation~\ref{eq:idec_qij}. One primary justification for t-distribution is its heavier tails compared to a Gaussian distribution. A heavier tail aids in efficiently mapping outliers observed in high-dimensional input data space to a lower 2D space for data visualization. 
\begin{equation}
\label{eq:idec_qij}
   q_{ij} =  \frac{(1 + || z_i - z_j ||)^{-1}}{\sum_{k\neq l}(1 + || z_k - z_l ||)^{-1}}
\end{equation}
Therefore, data points placed at a moderate distance in high-dimension are pulled farther by a t-distribution to aid data visualization in 2D space. In the context of cluster embedding, we argue that the additional separation between points in low dimensions may alter their cluster assignments. This phenomenon is observed in Figure~\ref{t-SNE-PCA} where high-dimensional deep image features are mapped on to 1) t-SNE and 2) two principal component spaces. The scattering of data points is evident in the t-SNE mapping (Figure~\ref{t-SNE-PCA} (a)), where one blue point appears on the left side of the figure, corrupting its cluster assignment, unlike the PCA mapping (Figure~\ref{t-SNE-PCA} (b)). This observation is in line with the contrats between data visualization and clustering presented in Table~\ref{literature}.

\begin{table*}[t]
\centering
\caption{Comparison between neighborhood embedding proposed for t-SNE data visualization~\citep{VanDerMaaten2008} and cluster embedding proposed in DEC~\citep{Xie2016} inspired by t-SNE. $\alpha$ = degrees of freedom of t-distribution, d = dimension of low-dimensional embedding. W represents the trainable parameter of an autoencoder.}
\scalebox{0.75}{
\begin{tabular}{lll}
\toprule
& t-SNE ~\citep{VanDerMaaten2008}& DEC~\citep{Xie2016} or IDEC~\citep{Guo2017} \\
\midrule
Purpose & Embedding for data visualization in d=2 & Embedding for clustering in d$>2$ \\
\midrule
Low-dimensional                      & Sampled from Gaussian      &Autoencoder \\
  embedding ($z_i$)         & with low $\sigma^2$      &latent space \\
 \midrule  
Distance or similarity     &    Between sample                         & Between point \& cluster         \\
measure      &  points   ($x_i$, $x_j$)            & centroid ($x_{i}$, $\mu_j$) \\  
  \midrule          
Embedding          & t-distribution,             &t-distribution,           \\
distribution ($q_{ij}$)       & $\alpha$ = 1               &$\alpha$ = 1            \\
                \midrule 
Target          & Gaussian in high-dimensional            &A function of            \\
distribution ($p_{ij}$)     & space (x)      &t-distributed $q_{ij}$        \\ 
  \midrule
Learning             & $z_{i+1} = z_i+ \frac{d~KLD (p, q)}{d(z_i)}$ &$ w_{i+1} = w_i+\frac{d~KLD (p, q)}{d(w)}$    \\
\bottomrule
\end{tabular}
\label{literature}
}
\end{table*}

\subsection {Embedding for clustering}
Embedding for clustering is achieved by infusing cluster separation information into the low-dimensional latent space of a deep neural network. 

While neighborhood embedding is initialized by sampling from a Gaussian distribution, cluster embedding methods use embedding learned from an autoencoder's latent space ($Z_i\in \Re^m$, where $m\ll d$). However, the current cluster embedding methods use the same t-distribution (Equation 4) to define the embedding distribution ($q_{ij}$), similar to neighborhood embedding. The target distribution ($p_{ij}$) is derived as a function of $q_{ij}$, as shown below.
\begin{eqnarray}
\label{eq:idec_pij}
s_{ij} =  \frac{q_{ij}^2}{\sum_{j} q_{ij}},~~ p_{ij} =  \frac{s_{ij}} 
{\sum_j s_{ij} }.
\end{eqnarray}
While pair-wise sample distances in neighborhood embedding have a complexity of O ($N^2$), the distances from the cluster centroids in embedding are O(N*K). Here, K is the number of clusters, much smaller than the number of samples (N). While an outlier point results in N large distances (extremely small $p_{ij}$ values) in neighborhood embedding, there will be much fewer (K$<<$N) of those large distances in cluster embedding. Therefore, the effect of outliers on cluster embedding can be much lower than the assumption made in neighborhood embedding for data visualization. 

Subsequently, the soft cluster assignment of the $i-th$ sample to the $j-th$ cluster  ($s_{ij}$) can be obtained using a Gaussian kernel as below, which is the negative exponent of the Mahalanobis distance ($d_{ij}$) (Equation \ref{eq:gceals_d_ij}) between the point ($Z_i$) and the $j$-th cluster centroid (Equation \ref{eq:gceals_s_ij}).
\begin{eqnarray}
\label{eq:gceals_d_ij}
d_{ij} &=& \sqrt  { (Z_i - \mu_j)~\Sigma_j^{-1}~(Z_i - \mu_j)^T } \\
\label{eq:gceals_s_ij}
s_{ij} &=& \exp~(- d_{ij})
\end{eqnarray}
The soft cluster assignments for each sample are then normalized to get cluster likelihood distribution, $p'_{ij}$ (Equation \ref{eq:gceals_prior_pij}). The cluster prior probability is initialized as $1/K$ as equally weighted clusters ($\omega_j$). Subsequently, ($\omega_j$) is updated from the second training epoch using $p'_{ij}$ estimated from the previous epoch, as shown below. 
\begin{eqnarray}
p'_{ij} = \frac{s_{ij}}{\sum_{j} s_{ij}}, ~~~ w_{j} = \frac{1}{N} \sum_{i=1}^N p'_{ij}  
\label{eq:gceals_prior_pij}
\end{eqnarray}
 
\section{Proposed method}  
We propose a novel deep clustering method, Gaussian Cluster Embedding in Autoencoder Latent Space (G-CEALS), as follows. First, a multivariate Gaussian distribution replaces the widely used t-distribution (Equation \ref{eq:idec_qij}). Unlike t-distribution, Gaussian distribution can regulate the cluster variance or scatter. Deep learning methods involving mixtures of Gaussian distributions have previously shown promising results in anomaly and error detection tasks~\citep{Zong2018dagmm,variani2015gaussianWordError}. Howeever, similar methods are not adjusted for the imbalances in multiple Gaussian distributions or clusters. 

\begin{figure*}[t]
\centering
 \includegraphics[trim=0 0.25cm 0 0.25cm, width=.65\textwidth] {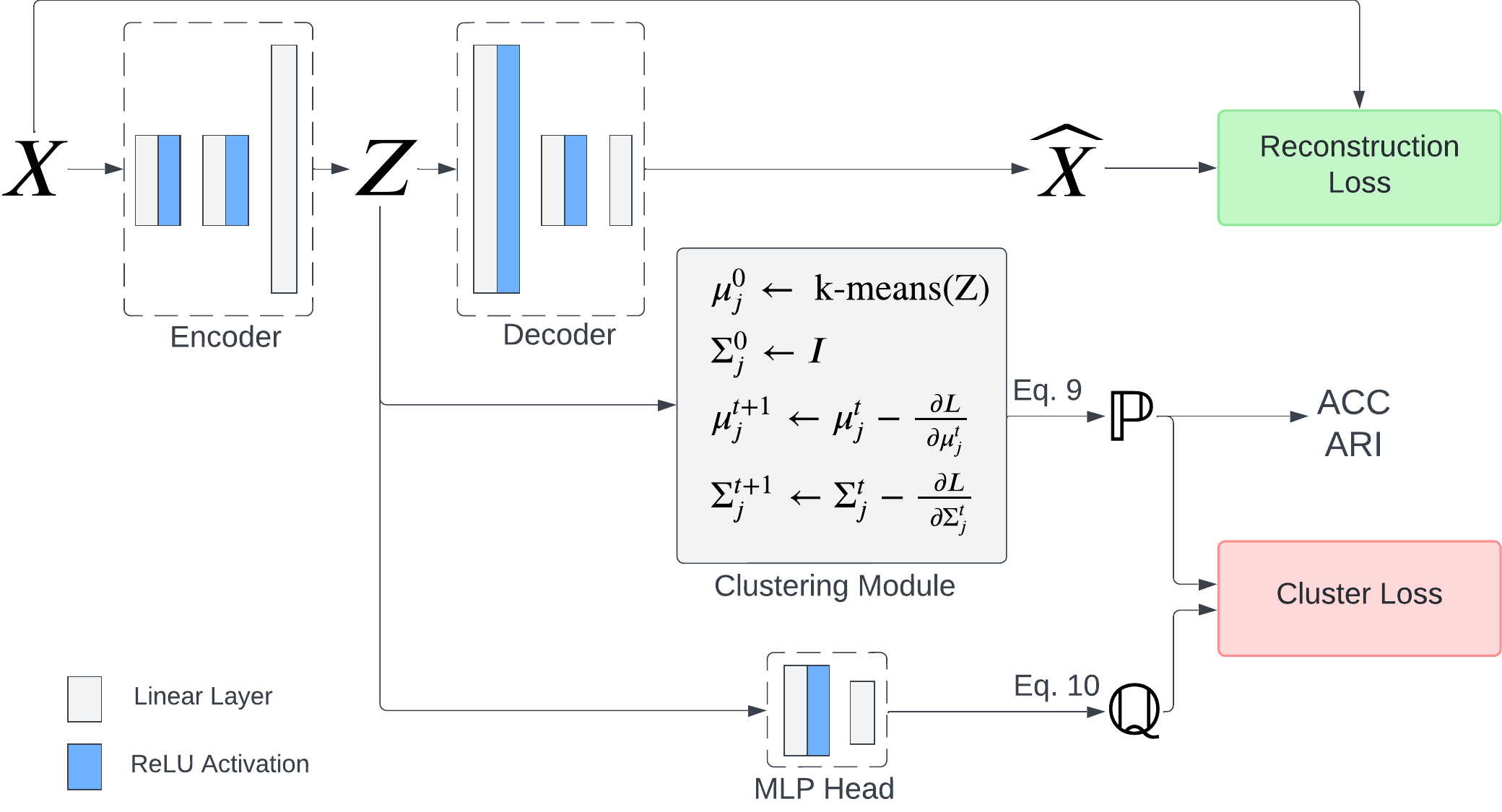}
 \caption{Proposed deep clustering framework for tabular data. All samples of an unlabeled tabular data set are used to train the autoencoder in tandem with two subnetworks: a clustering module and an MLP head with a softmax output layer. The final cluster distribution (P) and assignments are obtained after the clustering module. The final cluster assignments are evaluated using ACC, ARI, and NMI performance metrics.}
\label{evaluation}
\end{figure*}

The clustering on embedding ($Z_i \in \Re^m$) yields $K$ cluster assignments for individual samples. Each cluster $j$ is modeled by a centroid vector ($\mu_j \in \Re^m$) and a covariance matrix ($\Sigma_j \in \Re^{m\times m}$). Second, the proposed deep learning framework learns cluster distributions by updating $\mu_j$ and diagonals of $\Sigma_j$ as trainable parameters. Therefore, the deep learning framework learns Gaussian distributions on embedding space. The distribution parameters are initialized using K-means clustering means and by setting $\Sigma_j$ to an identity matrix, as shown in Figure~\ref{evaluation}.

We know from the Bayes' rule that the product of the likelihood distribution $p'_{ij}$ and prior probability $\omega_j$ is proportional to the posterior probability of a sample belonging to a cluster, as shown in Equation \ref{eq:gceals_pij}.
\begin{eqnarray}
\label{eq:gceals_pij} 
p_{ij} &=& \omega_{j} \times p'_{ij}
\end{eqnarray}
Additionally, a two-layer MLP head $f(\cdot)$ projects the input embedding space ($Z_{ij}$), which is transformed into soft cluster assignments $q_{ij}$ via a softmax layer, as shown in Equation \ref{eq:gceals_qij}.
\begin{eqnarray}
\label{eq:gceals_qij}
q_{ij} &=& \frac{\exp~(f(Z_{i,j}))}{\sum_{j} \exp~(f(Z_{i, j})}
\end{eqnarray}
The third important contribution of the proposed method is that the cluster (P) and target (Q) distributions are independently defined and dynamically updated, unlike existing embedding clustering methods.

\subsection {Optimizing deep embedding for clustering}

An autoencoder is trained to encode the input ($X\in \Re^{n \times m}$) into a latent embedding ($Z\in \Re^{n \times d}$), which is then decoded to reconstruct the original input ($\hat{X_i}$), as shown in the autoencoder's reconstruction loss below.
\begin{equation}
 \mathcal{L}_{recon} = \underset{\theta, \Phi}{\operatorname{argmin}} \frac{1}{N}\sum^{N}_{i=1} ||  X_i - \hat{X_i} ||_2^2.
 \label{eq:recon_loss}
\end{equation}
Here, $\theta$ and $\Phi$ denote the trainable parameters of the encoder and decoder, respectively. Notably, sophisticated autoencoder architectures (e.g., stacked, denoising, convolutional, variational) and popular learning tricks (e.g., data augmentation, dropout learning) may overshadow the contribution of a learning algorithm. Therefore, we use a standard autoencoder architecture without data augmentation and dropout learning. Similar autoencoders have been found effective in the supervised classification of tabular data~\citep{abrar2022perturbation, rabbani2024attention}. The pretrained autoencoder network is subsequently fine-tuned to yield cluster-friendly embedding and cluster assignments. A clustering module with trainable parameters ($\mu$ and $\Sigma$) computes the cluster distribution ($P$) using Equation \ref{eq:gceals_pij}. A cross-entropy loss involving $P$ and $Q$ distributions is minimized as the clustering loss in Equation \ref{eq:cluster_loss}. The cross-entropy loss induces separation between clusters instead of collapsing to cluster centroids.
\begin{eqnarray}
\label{eq:cluster_loss}
    \mathcal{L}_{cluster} &=& \frac{1}{N} \sum_{i=1}^N \sum_{j=1}^K p_{ij} \log{q_{ij}} 
\end{eqnarray}
The pretrained autoencoder is jointly trained with the clustering loss to learn cluster distributions in the latent space using Equation \ref{eq:gceals_objective}.
\begin {eqnarray}
\label{eq:gceals_objective}
\mathcal{L} &=& \mathcal{L}_{recon} + \gamma* \mathcal{L}_{cluster}  \\ 
&=& \underset{\theta, \Phi, \mu, \Sigma}{\operatorname{argmin}} \frac{1}{N} \sum^{N}_{i=1} ||  X_i - \hat{X_i} ||_2^2 + \gamma * \frac{1}{N} \sum_{i=1}^N \sum_{j=1}^K  p_{ij} \log{q_{ij}}
\nonumber
\end {eqnarray}
Here, $\gamma$ is a hyperparameter to balance the contribution of clustering loss in joint learning. The effect of $\gamma$ values on the clustering accuracy is evaluated later in an ablation study. The computational steps of the proposed framework are presented in Algorithm 1.
\begin{algorithm}[t]
\caption{Proposed G-CEALS Deep Clustering Algorithm}
\begin{algorithmic}[1]
\State Input: $d$-dimensional tabular data, $X\in\Re^{n\times d}$
\State Output: Cluster-friendly embedding, $Z\in\Re^{n\times m}$, m$\ll$d and soft cluster assignments ($q_{ij}$)
\State Pre-train autoencoder ($\{ W_{encoder}$, $W_{decoder}\}$) $\leftarrow$ X
\State Embedding ($Z$) $\leftarrow$ Encoder ($X$, $W_{encoder}$)
\State Initialize pseudo-labels: $\Tilde{Y} \leftarrow$ k-means($Z$) 
\State Initialize $j$-th cluster parameters: $\mu_j \leftarrow$ k-means($Z$), $\Sigma_j \leftarrow I, \omega_j \leftarrow 1/k $
\State Trainable cluster distribution parameters: $W_{cluster}$ $\leftarrow$ $\{[\mu_1, \mu_2, ..., \mu_k], [\Sigma_1, \Sigma_2, ..., \Sigma_k]$ $\}$   
\State Initialize: $W^0$ =$\{ W_{encoder}$, $W_{decoder}$, $W_{cluster}$, $W_{MLP}$ $\}$ 
\For {t = 1 $\rightarrow$ n\_epochs}
\State $X_b$ $\leftarrow$ Sample mini-batch from $X$ for uniform class distribution
\State $Z^t$ $\leftarrow$ Encoder ($X_b$, $W_{encoder}^t$), $\hat{X_b}$ $\leftarrow$ Decoder ($Z^t$, $W_{decoder}^t$)
\State $p_{ij}$ $\leftarrow$ ($Z^t$, $W^t_{cluster}$) using Eq. \ref{eq:gceals_pij}
\State $q_{ij}$ $\leftarrow$ ($Z^t$, $W^t_{MLP}$) using Eq. \ref{eq:gceals_qij} 
\State $\mathcal{L}$ $\leftarrow$ $\mathcal{L}_{recon} + \gamma* \mathcal{L}_{cluster}$, Eq. \ref{eq:gceals_objective} 
\State $W^{t+1}$ $\leftarrow$ $W^{t}$ - $\alpha$$\nabla_{W^t}$$\mathcal{L}$, update trainable parameters minimizing the joint loss in Eq. \ref{eq:gceals_objective}
\State $p'_{ij}$ $\leftarrow$ ($Z$, $W^{t+1}_{cluster}$), $Z$ $\leftarrow$ Encoder ($X$, $W_{encoder}^{t+1}$) using Eq. \ref{eq:gceals_prior_pij}
\State $\omega_{j}$ $\leftarrow$ update using Eq. \ref{eq:gceals_prior_pij} 
\If{$\omega_j \le 1/2k$}
\State Stop training 
\EndIf
\EndFor
\end{algorithmic}
\end{algorithm}

\subsection {Cluster imbalance and convergence}

The convergence of the proposed cluster loss is important in addition to ensuring the cluster separation after training. Figure~\ref{fig:loss_curves} shows smooth convergence of clustering loss for two different $\gamma$ values. A larger $\gamma$ value (1.0) speeds up the convergence. However, a lower $\gamma$ value (0.1) helps with stable and smooth convergence at a slower pace. Therefore, a $\gamma$  value of 0.1 is chosen for models that require this hyperparameter. 

The effectiveness of the proposed deep clustering method in creating cluster separation is visualized using t-SNE plots in Figure~\ref{fig:tsne}. The cluster visualization identifies an issue where minority clusters may merge with majority ones after training for a long period without an early stopping. The merging of clusters happens due to cluster imbalance in tabular data sets. We address this issue by adopting two strategies. First, we implement an early stopping criterion based on the cluster weight $\omega_j$ updates, which is a measure of cluster size. The cluster weight becomes zero when it merges with another cluster during training, which can be prevented by setting a threshold on cluster weight. When there are K balanced clusters, the weights initially take a value of 1/K. We stop the training when at least one of the cluster weights drops below 50\% of 1/K to prevent possible cluster merging.

Second, we use mini-batch gradient descent to optimize the deep clustering model. However,  mini-batches may not contain all cluster samples when the data set is imbalanced, leading to inflated or biased clustering accuracy. We use K-means clustering to obtain the pseudo-labels and identify the minority cluster. If the minority cluster has $n_{min}$ samples, we randomly choose an equal number of samples from other clusters to form a batch size of 256 or lower. This random sampling is performed at every epoch and repeated 1000 times to train the model. Therefore, the batch size varies across the data sets depending on the size of the minority cluster.

The convergence of three cluster parameters: mean vectors, covariance matrices, and cluster weights is presented in Figure~\ref{fig:parameter_curves} for a two-cluster clustering problem. For better visualization, we use the L2 norm distance between two consecutive mean vector updates and the determinant of the covariance matrices.

\begin{table}[t]
\centering
\caption{Summary of sixteen tabular data sets used in this study. The \emph {feature dimension} combines numerical and one-hot encoded categorical features. F-S ratio is the ratio of features to samples. C-scores represent the mean absolute correlations across all feature pairs.}
\label{tab:datasets}
\scalebox{0.75}{
\begin{tabular}{llrrrrrrrrr}
\toprule
OpenML & Name & Samples & Numerical & Categorical & Feature dimension & Classes & F-S ratio & C-score \\
\midrule
1063 & Kc2 & 522 & 21 & 0 & 21 & 2 & 4.023 & 0.557 \\
40994 & Climate-model-simulation-crashes & 540 & 20 & 0 & 18 & 2 & 3.333 & 0.000 \\
1510 & Wdbc & 569 & 30 & 0 & 30 & 2 & 5.272 & 0.101 \\
1480 & Ilpd & 583 & 9 & 1 & 11 & 2 & 1.887 & 0.036 \\
11 & Balance-scale & 625 & 4 & 0 & 4 & 3 & 0.640 & 0.000 \\
37 & Diabetes & 768 & 8 & 0 & 8 & 2 & 1.042 & 0.000 \\
469 & Analcatdata\_dmft & 797 & 0 & 4 & 21 & 6 & 2.635 & 0.005 \\
458 & Analcatdata\_authorship & 841 & 70 & 0 & 70 & 4 & 8.323 & 0.000 \\
1464 & Blood-transfusion-service-center & 748 & 4 & 0 & 4 & 2 & 0.535 & 0.167 \\
1068 & Pc1 & 1109 & 21 & 0 & 21 & 2 & 1.894 & 0.271 \\
1049 & Pc4 & 1458 & 37 & 0 & 37 & 2 & 2.538 & 0.063 \\
23 & Cmc & 1473 & 2 & 7 & 24 & 3 & 1.629 & 0.011 \\
1050 & Pc3 & 1563 & 37 & 0 & 37 & 2 & 2.367 & 0.176 \\
40975 & Car & 1728 & 0 & 6 & 21 & 4 & 1.215 & 0.000 \\
40982 & Steel-plates-fault & 1941 & 27 & 0 & 27 & 7 & 1.391 & 0.046 \\
1067 & Kc1 & 2109 & 21 & 0 & 21 & 2 & 0.996 & 0.481 \\
\bottomrule
\end{tabular}
}
\end{table}

\section {Experiments}

This section identifies the tabular data sets, baseline algorithms, and metrics used to evaluate the performance of our proposed deep clustering method.

\subsection{Tabular data sets}


All methods are evaluated on a diverse set of 16 tabular datasets sourced from OpenML-CC18~\citep{OpenMLCC18}. Table~\ref{tab:datasets} summarizes 16 tabular data sets representing various application domains and a wide range of data statistics. The heterogeneity of tabular data sets is further characterized by F-S ratio and C-scores in the Table. FS-ratio represents the feature and sample ratio of the data set. The C-score provides a measure of feature correlations. It shows the mean of the absolute correlations across all features. It is notable that prior studies on tabular data classification (not clustering) selectively use data sets with very large sample sizes~\citep{Arik2021Tabnet, Kossen2021Npt, Gorishniy2021Ftt}. In practice, most tabular data domains include limited samples and features with or without the presence of categorical variables, unlike image data sets.

\subsection{Adapting baseline methods to tabular data}
Recent surveys on deep clustering methods show no examples of clustering tabular data sets~\citep{zhou2022comprehensive, Ren2022survey}. Deep embedding clustering methods have been invariably designed and evaluated on benchmark image data sets. Therefore, existing deep clustering methods may not be ideal baselines for tabular data due to the data-centric contrasts presented in Table~\ref{imageVStable}. 
 
 The DEC~\citep{Xie2016} and IDEC~\citep{Guo2017} methods use a fully-connected autoencoder architecture as d–500–500–2000–10-2000-500-500-d. The deep k-means (DKM)~\citep{MoradiFard2020} and AE-CM~\citep{Boubekki2021} methods use the same learning architecture after replacing the fixed dimensionality of the embedding (10) with the number of target clusters (k). The dynamic autoencoder (DynAE) uses the same architecture as DEC/IDEC~\citep{Mrabah_neunet_2020}. However, the objective function is regularized by image augmentation (shifting and rotation), which has to be disabled for tabular data in this paper. Several other methods are built on convolutional neural network (CNN) architectures~\citep{Huang2020, Dizaji2017}, whereas fully connected neural networks are the default choice for tabular data. For example, Caron et al. have learned visual features from images using AlexNet and VGG-16 after Sobel filtering for color removal and contrast enhancement~\citep{Caron2018}, which are not applicable to tabular data. Their \emph{deepCluster} architecture has five convolutional layers with up to 384 2D image filters to learn image features. The transfer learning of tabular data using the image-pre-trained VGG-16 model is not trivial. The DEPICT method uses a convolutional denoising autoencoder for reconstructing original images from corrupted images~\citep{Dizaji2017}. Instead, we use a standard convolutional autoencoder replacing the 2D filters with 1D kernels to learn embedding for tabular data vectors because image denoising is not reproducible on tabular data.

 All these methodological aspects are considered in adapting seven state-of-the-art deep clustering methods (DEC, IDEC, AE-CM, DynAE, DEPICT, DKM, DCN) as baseline deep clustering methods for tabular data sets, including two traditional clustering methods (k-means) and Gaussian Mixture Model (GMM).

\subsection {Evaluation}
The proposed deep clustering model training entails self-supervised data reconstruction and cluster distribution learning without involving ground truth labels.  The quality of cluster embedding is evaluated in downstream clustering using clustering accuracy (ACC)~\citep{Kuhn1955}, as shown in Equation~\ref{equation-LLE2}.

\begin{equation}
ACC  = \underset{m}{\operatorname{max}} \frac{\sum^{N}_{i=1} 1 \{y_{true}(i) = m(y_{pred}(i))\}}{N}
 \label{equation-LLE2}
\end{equation}
Here, $y_{true}$(i) is the ground truth class label of the i-th sample. $y_{pred}$(i) is the predicted cluster label. m() finds the best label mapping between the cluster and ground truth labels. The mapping can be obtained by the Hungarian algorithm~\citep{Kuhn1955}. Notably, a cross-validation of the model is not considered in unsupervised learning where class labels are not involved in the training process.  Therefore, we use all data samples in model training, similar to training and evaluating any clustering algorithms~\citep{Xie2016,Guo2017}. The clustering accuracy takes a value between 0 (failure) and 1 (perfect clusters). In all experiments, the cluster number is set to the number of class labels in a given data set. The accuracy scores are multiplied by 100 to represent the numbers in percentage.

Likewise, Adjusted Rand Index (ARI) quantifies the similarity between the true and predicted clusters while considering chance, as shown in Eq.~\ref{eq:eval_ari}. ARI adjusts for randomness or the expected similarity due to random chance, as shown below. ARI is particularly useful when the number of clusters is not known in advance. Its value ranges between -0.5 and +1. ARI scores close to 0.0 indicate random cluster labels, and a higher positive value suggests better concordance between the true and predicted clusters.
\begin{equation}
\label{eq:eval_ari}
     ARI = \frac{RI - \mathbb{E}(RI)}{max(RI) - \mathbb{E}(RI)}, ~~~RI = \frac{TP + TN}{{}^{N}C_2}.
\end{equation}
Here, TP and TN denote the true positive and true negative pairs, ${}^{N}C_2$ denotes the number of possible pairs, and $\mathbb{E}(RI)$ is the expected Rand index. For multiclass classification, TP and TN are determined using the one-vs-all scheme.

\section {Results}

All experiments are conducted on a Dell Precision 5820 workstation running Ubuntu 20.04 with 64GB RAM and an NVIDIA Quadro RTX A5000 GPU with 16GB memory. We standardize numerical features using their mean and standard deviation and one-hot encode categorical features before model training. The source code is publicly available on github\footnote{\href{https://github.com/mdsamad001/G-CEALS---Deep-Clustering-for-Tabular-Data/}{https://github.com/mdsamad001/G-CEALS---Deep-Clustering-for-Tabular-Data/}}.

\subsection {Learning architecture and implementation}
All algorithms are implemented in Python. The proposed deep learning method is developed using the PyTorch package, whereas traditional methods are implemented using scikit-learn. The baseline implementations are obtained from their respective GitHub repositories. Specifically, the DEPICT algorithm is implemented using the Theano package, and other methods are implemented using the TensorFlow or Keras package. 

As mentioned in Section 5.2, all baseline deep clustering algorithms are benchmarked on image datasets. Therefore, minimal modifications are made to the source code to enable the input and learning of tabular data sets in place of image data.
All methods, including the proposed one, use a fully connected autoencoder of the same architecture (d-500-500-2000-m-2000-500-500-d), where $m$ is the embedding dimension. The DKM method presets the value $m$ to the number of clusters (k). The adapted DEPICT method uses a CNN-based architecture with 1D filters for tabular data. For all experiments, the learning rate is set to 0.001 with an Adam optimizer and using a batch size of 256. Each method pretrains an autoencoder for 1000 epochs and then fine-tunes jointly with the clustering loss (Eq. \ref{eq:gceals_objective}) for another 1000 epochs.

An obvious benefit of deep learning methods over traditional clustering is its flexible embedding size. Therefore, the embedding dimension is varied from 5 to 20 at five intervals (5, 10, 15, and 20) for each deep learning method and data set pair, considering the heterogeneity in tabular data sets and features. However, the DMK method sets the embedding size equal to the number of clusters. The deep clustering methods are compared on the embedding dimension that yields the best clustering performance for a given data set.

\begin{figure*}[t]
\vspace{-10pt}
\subfigure[$\gamma = 0.1$] { \includegraphics[trim=0 .33cm 0 0cm, clip, width=0.36\textwidth] {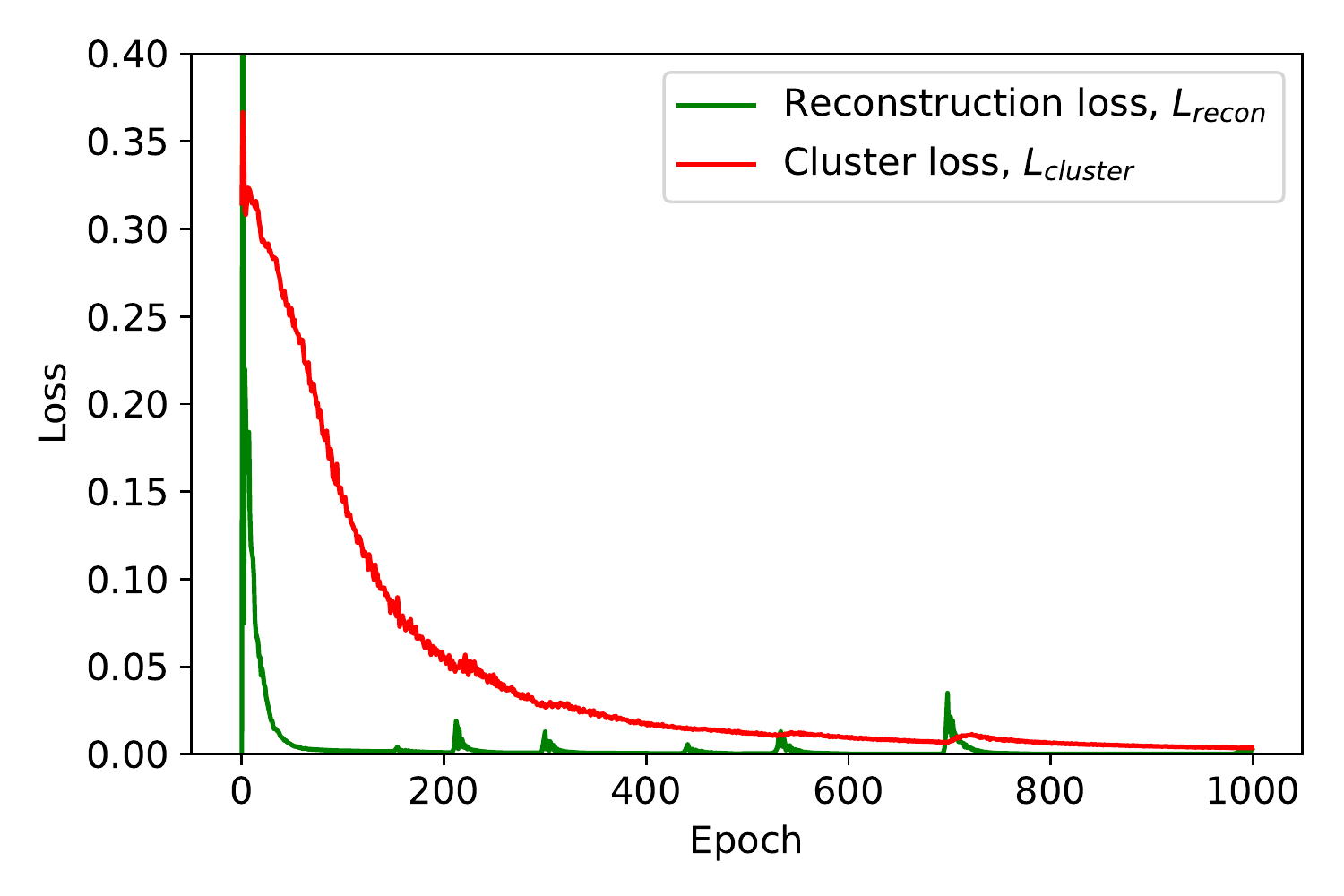}}
\hspace{-10pt}
\subfigure[$\gamma = 1.0$] { \includegraphics[trim=0 .33cm 0 0cm, clip, width=0.36\textwidth] {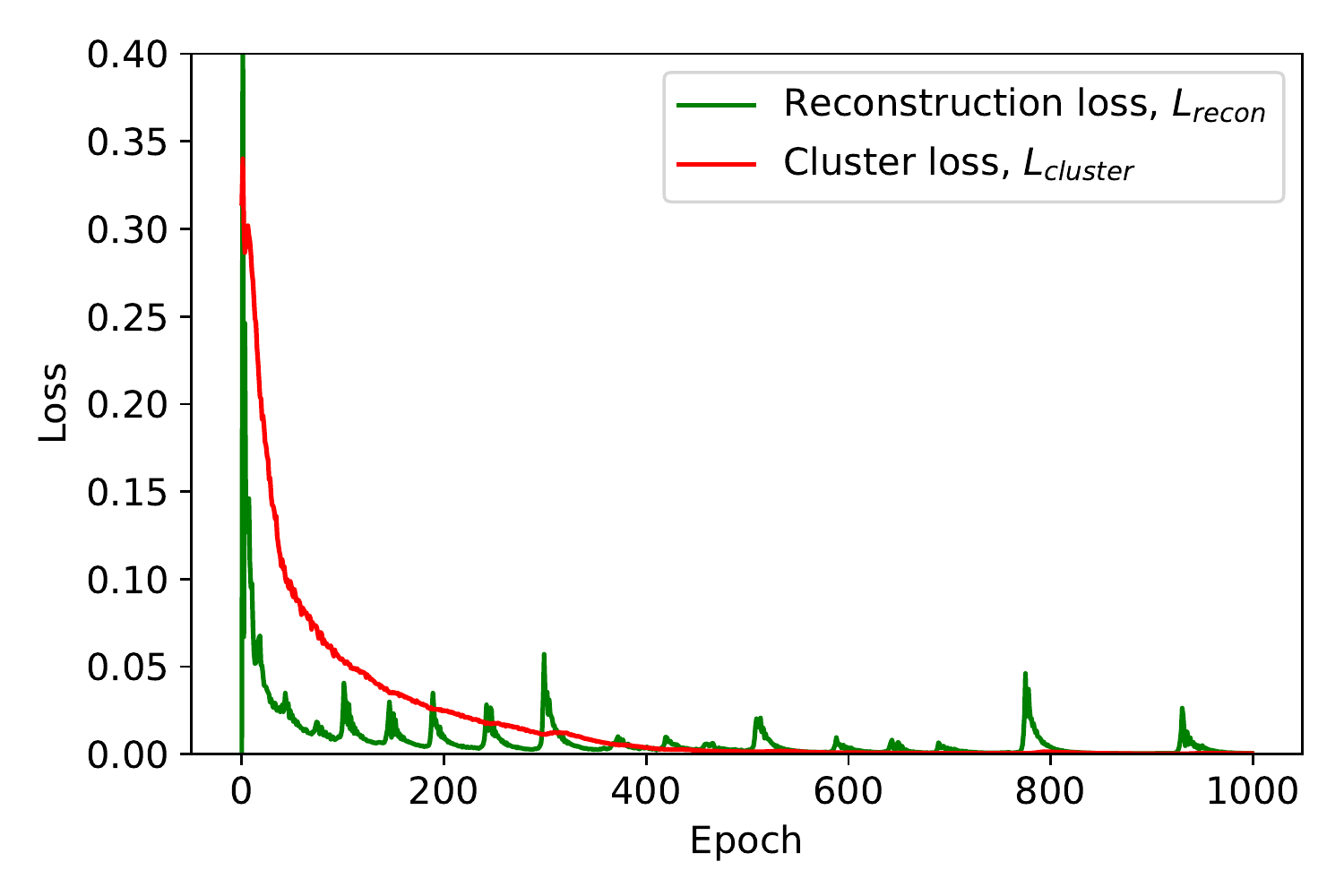}}
\vspace{-10pt}
\caption{The reconstruction and clustering losses are obtained using the tabular data set with ID 1510 for two $\gamma$ values. A higher $\gamma$ value results in faster convergence of the clustering loss, slowing the reconstruction loss. However, a lower value is preferred to ensure smooth convergence of the cluster parameters and autoencoder weights.}
\label{fig:loss_curves}
\end{figure*}

\subsection {Cluster imbalance and convergence}

The convergence of the proposed cluster loss is important in addition to ensuring the cluster separation after training. Figure~\ref{fig:loss_curves} shows smooth convergence of clustering loss for two different $\gamma$ values. A larger $\gamma$ value (1.0) speeds up the convergence. However, a lower $\gamma$ value (0.1) helps with stable and smooth convergence at a slower pace. Therefore, a $\gamma$  value of 0.1 is chosen for models that require this hyperparameter. 

\begin{figure*}[t]
\vspace{-10pt}
\subfigure[Before clustering] { \includegraphics[trim=0 0 0 0cm, clip, width=0.33\textwidth] {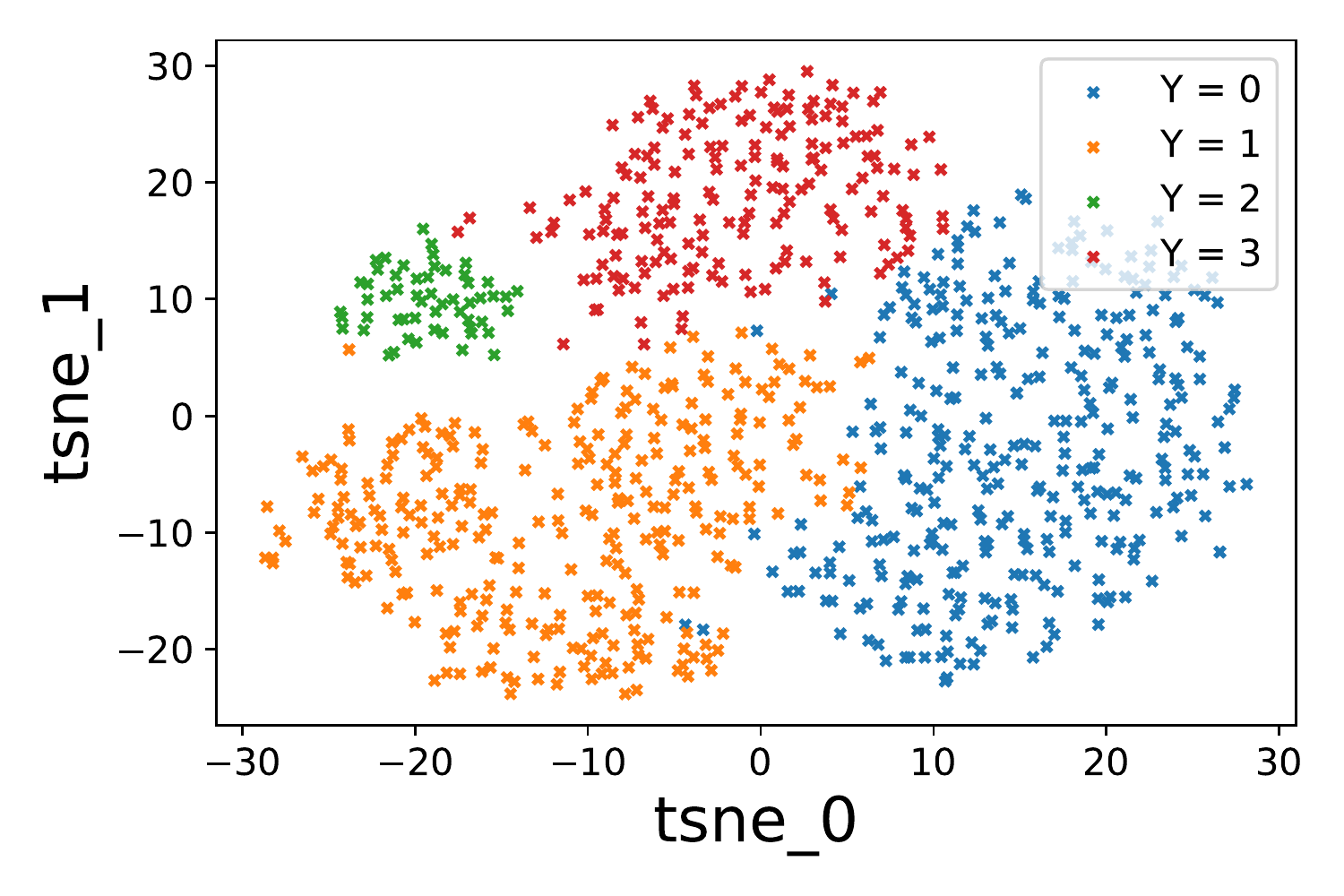}}
\hspace{-10pt}
\subfigure[Clustering without early stop] { \includegraphics[trim=0 0 0 0cm, clip, width=0.33\textwidth] {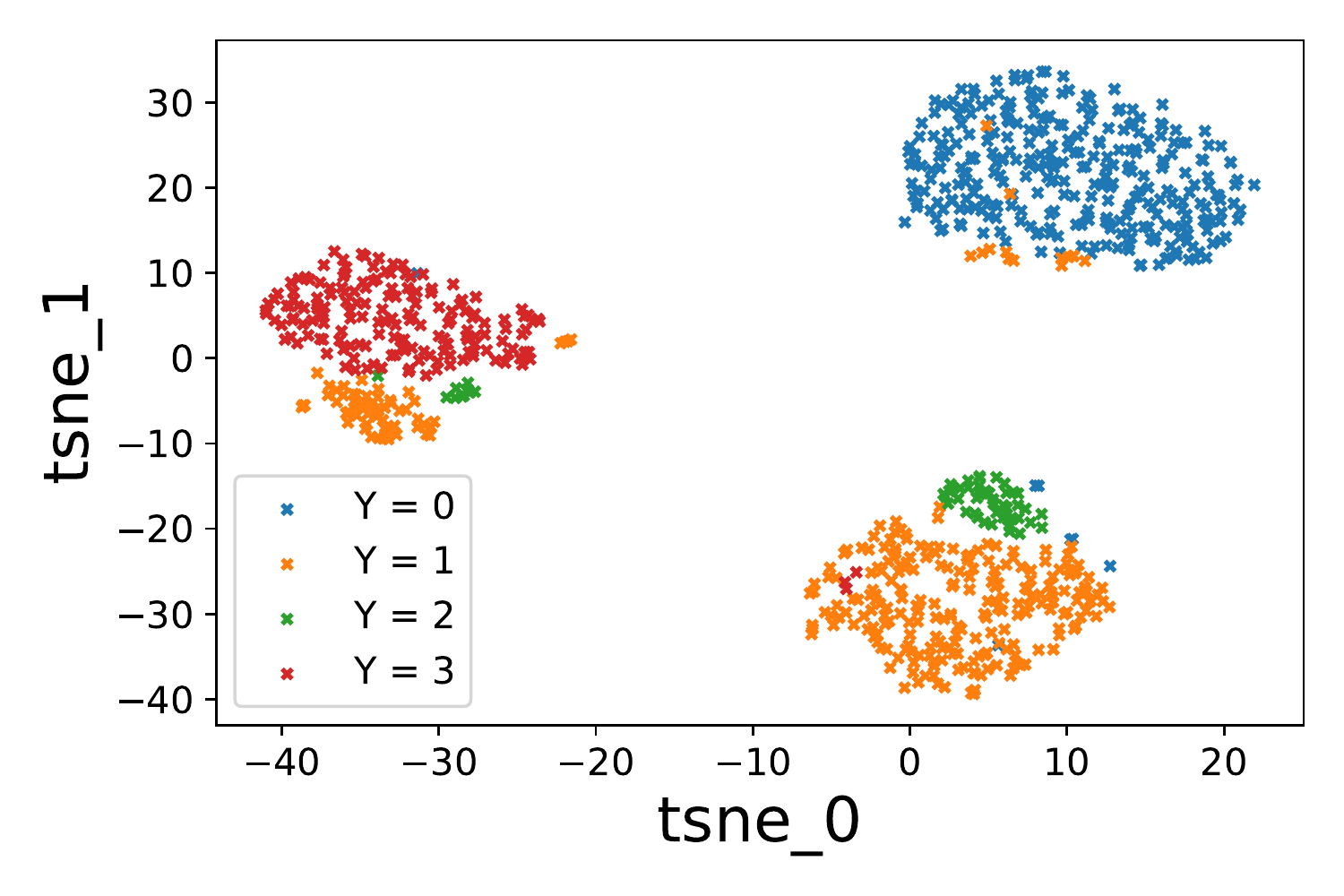}}
\hspace{-10pt}
\subfigure[Clustering with early stop] { \includegraphics[trim=0 0 0 0cm, clip, width=0.33\textwidth] {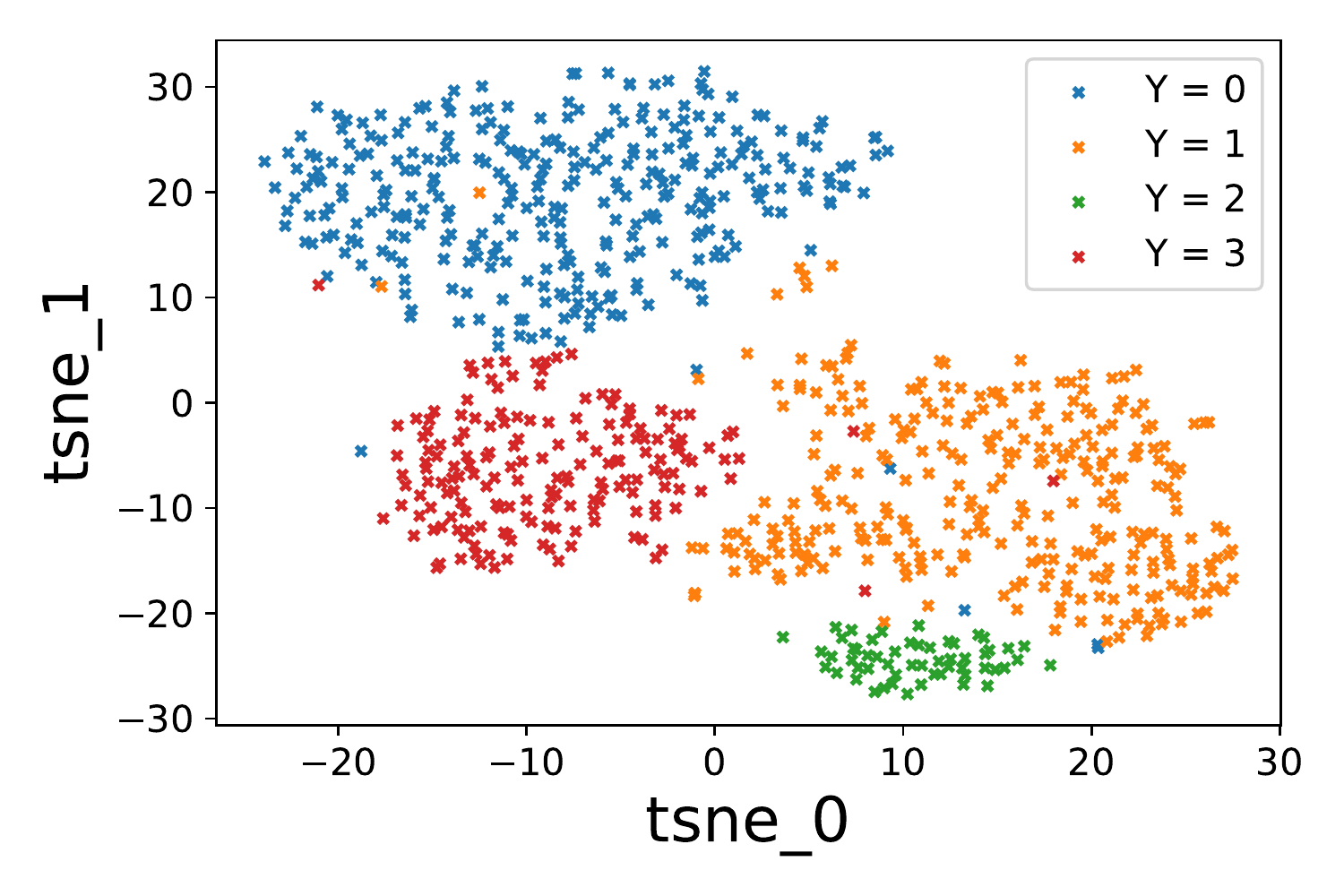}}
\vspace{-10pt}
\caption{t-SNE visualization of deep clustering of data set ID 458 with and without early stopping. While the deep clustering method facilitates cluster separation, it may merge the minority clusters due to the cluster imbalance in tabular data.}
\label{fig:tsne}
\end{figure*}

The effectiveness of the proposed deep clustering method in creating cluster separation is visualized using t-SNE plots in Figure~\ref{fig:tsne}. The cluster visualization identifies an issue where minority clusters may merge with majority ones after training for a long period without an early stopping. The merging of clusters happens due to cluster imbalance in tabular data sets. We address this issue by adopting two strategies. First, we implement an early stopping criterion based on the cluster weight $\omega_j$ updates, which is a measure of cluster size. The cluster weight becomes zero when it merges with another cluster during training, which can be prevented by setting a threshold on cluster weight. When there are K balanced clusters, the weights initially take a value of 1/K. We stop the training when at least one of the cluster weights drops below 50\% of 1/K to prevent possible cluster merging.

Second, we use mini-batch gradient descent to optimize the deep clustering model. However,  mini-batches may not contain all cluster samples when the data set is imbalanced, leading to inflated or biased clustering accuracy. We use K-means clustering to obtain the pseudo-labels and identify the minority cluster. If the minority cluster has $n_{min}$ samples, we randomly choose an equal number of samples from other clusters to form a batch size of 256 or lower. This random sampling is performed at every epoch and repeated 1000 times to train the model. Therefore, the batch size varies across the data sets depending on the size of the minority cluster.

\begin{figure*}[t]
\vspace{-15pt}
\subfigure[Cluster centroid updates] { \includegraphics[trim=0 0 0 0cm, clip, width=0.33\textwidth] {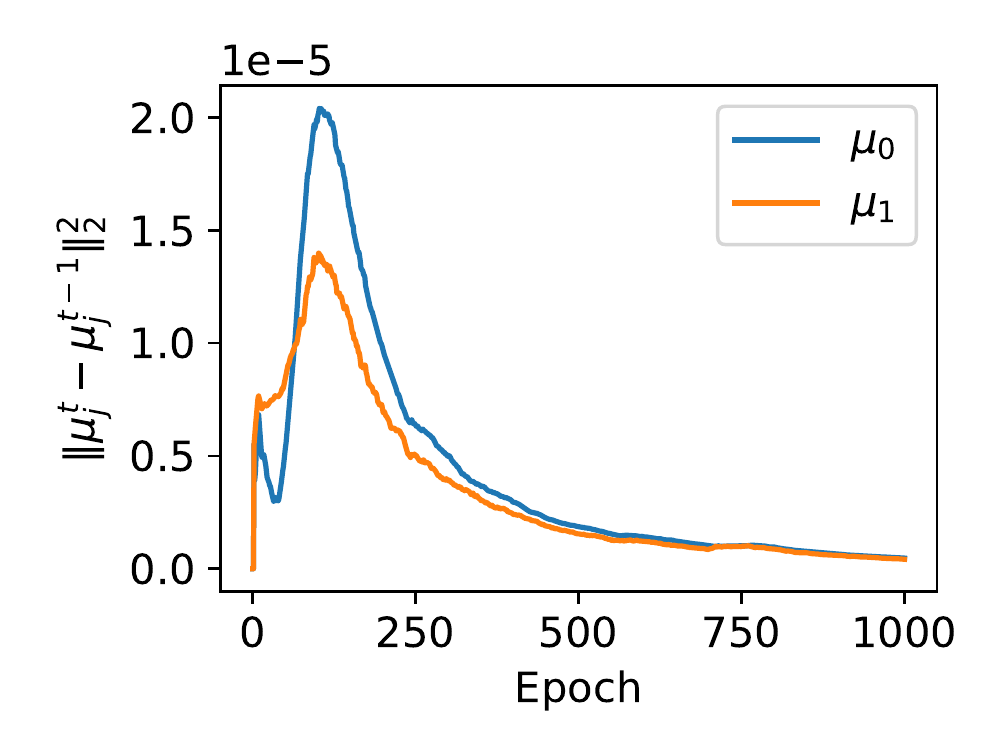}}
\hspace{-10pt}
\subfigure[Determinant of cluster covariances] { \includegraphics[trim=0 0 0 0cm, clip, width=0.33\textwidth] {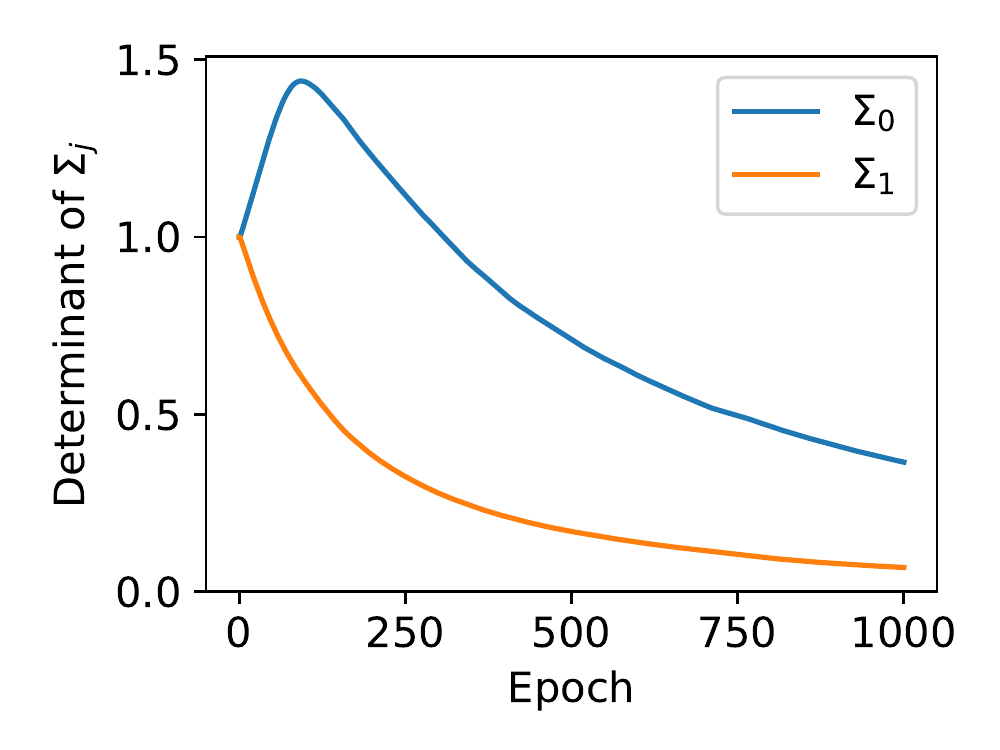}}
\hspace{-10pt}
\subfigure[Cluster weights] { \includegraphics[trim=0 0 0 0cm, clip, width=0.33\textwidth] {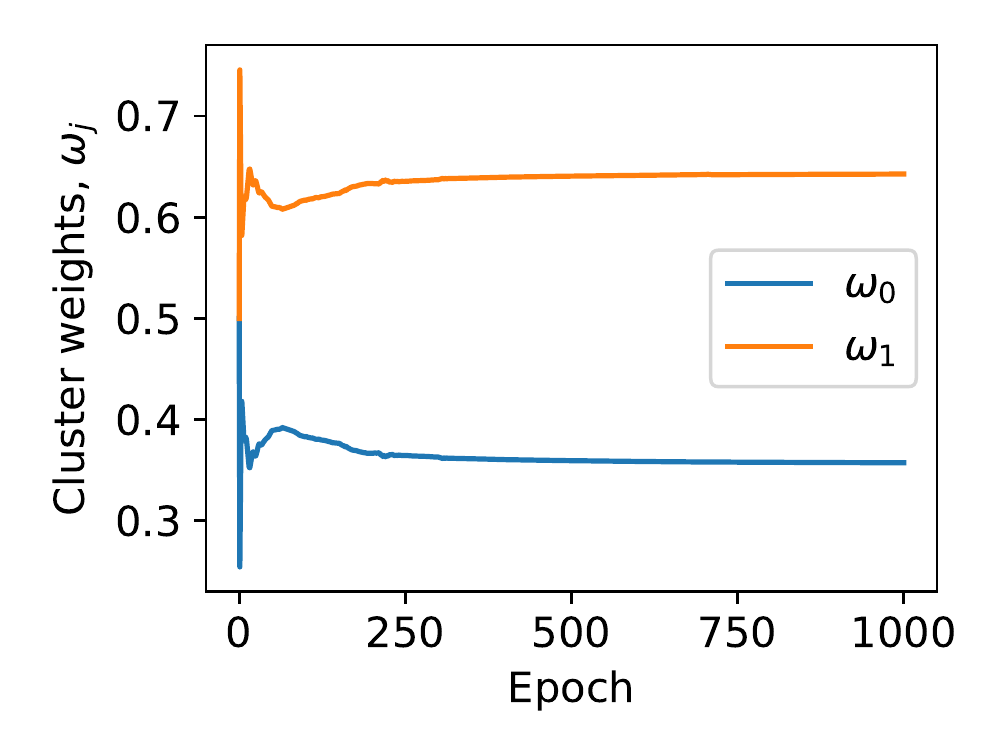}}
\vspace{-10pt}
\caption{Convergence of cluster centroids ($\mu_j$), cluster covariances ($\Sigma_j$), and cluster weights ($\omega_j$) for two clusters using dataset ID 1510. Here, $t$ represents epoch, and $j$ is the cluster index.
}
\label{fig:parameter_curves}
\end{figure*}

The convergence of three cluster parameters: mean vectors, covariance matrices, and cluster weights is presented in Figure~\ref{fig:parameter_curves} for a two-cluster clustering problem. For better visualization, we use the L2 norm distance between two consecutive mean vector updates and the determinant of the covariance matrices.

\subsection{Embedding dimension for clustering}

Images are high-dimensional data and are conventionally projected onto a low-dimensional embedding for effective separation of class or clusters. Unlike image data, the feature dimension of tabular data can be considerably low and heterogeneous. In the absence of an effective feature extractor (like CNN used for images), it is unknown if a tabular data set can be better clustered at a higher or lower dimension than its original feature space. Therefore, the optimal embedding size for clustering tabular data may vary across data sets or deep clustering methods. We identify the embedding dimension that yields the best clustering performance for a given deep clustering method and tabular data set. We observe that low-dimensional tabular data sets can benefit from learning a higher dimensional embedding using an overcomplete autoencoder (latent space larger than the input space). Therefore, we use these best embedding dimensions to compare our proposed and baseline deep clustering methods in subsequent sections.

\newcolumntype{Y}{p{1.4cm}}
\begin{table}[h]
\caption{Comparison of clustering accuracy between the proposed G-CEALS method and baseline traditional or deep clustering methods on sixteen tabular data sets. }
\label{tab:results_with_acc}
\centering
\scalebox{0.65}{
\begin{tabular}{lcccccccccccc}
\toprule
Dataset ID & K-means(X) & GMM(X) & K-means(Z) & GMM(Z) & DEC & IDEC & DEPICT & DynAE & DCN & AE-CM & DKM & G-CEALS \\
\midrule
1063 & 80.3 & 80.3 & 68.0 & 72.8 & 67.0 & 66.9 & 66.1 & 79.5 & 80.3 & 79.6 & 79.6 & \bfseries 81.0 \\
40994 & 52.4 & 53.1 & 57.0 & 61.1 & 57.0 & 54.8 & 50.4 & 60.4 & 64.6 & \bfseries 91.5 & 60.0 & 85.9 \\
1510 & 91.2 & 94.0 & 92.1 & 89.5 & 91.4 & 93.1 & 93.7 & 58.0 & 77.7 & 62.7 & 80.2 & \bfseries 95.6 \\
1480 & 61.4 & 53.5 & 64.0 & 64.0 & 59.3 & 58.7 & 62.1 & 64.0 & 63.7 & \bfseries 71.3 & 65.8 & 65.2 \\
11 & 51.2 & 52.6 & 52.3 & 55.8 & 51.4 & 52.3 & 52.8 & 51.7 & 57.6 & 46.1 & 47.6 & \bfseries 69.3 \\
37 & 66.1 & 58.5 & 73.3 & 72.8 & 71.4 & 71.2 & 67.4 & 67.4 & 64.4 & 65.1 & 66.4 & \bfseries 74.1 \\
469 & 21.8 & 21.5 & 21.3 & 21.6 & 21.5 & 21.3 & 21.0 & 22.3 & \bfseries 22.6 & 20.5 & 21.9 & 22.0 \\
458 & 98.2 & 98.5 & 83.6 & 87.4 & 81.5 & 83.1 & 78.8 & \bfseries 99.9 & 51.0 & 49.5 & 64.5 & 89.8 \\
1464 & 68.3 & 57.6 & 73.9 & 69.1 & 66.4 & 66.4 & 56.1 & 60.0 & 74.1 & \bfseries 76.2 & 64.3 & 74.1 \\
1068 & 89.2 & 71.0 & 92.3 & 86.8 & 82.5 & 81.2 & 59.0 & 93.0 & 93.0 & \bfseries 93.1 & \bfseries 93.1 & 92.5 \\
1049 & 81.9 & 72.2 & 84.3 & 72.2 & 70.0 & 69.3 & 66.5 & 70.5 & 87.1 & \bfseries 87.9 & 87.5 & 85.3 \\
23 & 42.4 & 39.6 & 42.0 & 43.4 & \bfseries 44.3 & 43.8 & 42.0 & 38.8 & 40.7 & 42.7 & 36.9 & 44.2 \\
1050 & 87.3 & 75.0 & 66.3 & 76.2 & 58.6 & 62.5 & 57.9 & \bfseries 89.7 & 89.0 & \bfseries 89.7 & 64.9 & 83.0 \\
40975 & 31.2 & 31.2 & 31.1 & 34.7 & 31.1 & 31.1 & 38.3 & 46.1 & 56.5 & \bfseries 70.0 & 36.1 & 49.2 \\
40982 & 39.0 & 37.5 & 45.2 & 46.1 & 44.4 & 44.8 & 36.6 & \bfseries 47.6 & 38.0 & 36.6 & 44.9 & 41.3 \\
1067 & 83.3 & 72.6 & \bfseries 85.3 & 77.0 & 80.1 & 79.3 & 62.1 & 77.3 & 84.7 & 84.7 & \bfseries 85.3 & 85.2 \\
\midrule
Avg. Rank & 6.7 (2.5) & 7.9 (3.6) & 5.8 (2.6) & 5.6 (2.2) & 7.6 (2.6) & 7.9 (2.7) & 9.2 (3.3) & 5.8 (3.7) & 5.1 (3.7) & 5.6 (4.8) & 6.2 (3.6) & \bfseries 2.9 (1.7) \\ 
\midrule
Overall Rank & 8 & 11 & 5 & 3 & 9 & 10 & 12 & 6 & 2 & 4 & 7 & \bfseries 1 \\
\bottomrule
\end{tabular}
}
\end{table}
\begin{table}[t]
\caption{Comparison of adjusted Rand index (ARI) scores between the proposed G-CEALS and baseline traditional or deep clustering methods on sixteen tabular data sets.}
\label{tab:results_with_ari}
\centering
\scalebox{0.65}{
\begin{tabular}{lcccccccccccc}
\toprule
Dataset ID & K-means(X) & GMM(X) & K-means(Z) & GMM(Z) & DEC & IDEC & DEPICT & DynAE & DCN & AE-CM & DKM & G-CEALS \\
\midrule
1063 & 0.044 & 0.044 & 0.127 & 0.193 & 0.115 & 0.112 & 0.102 & 0.215 & 0.192 & 0.082 & 0.007 & \bfseries 0.335 \\
40994 & 0.001 & 0.003 & 0.017 & \bfseries 0.031 & 0.017 & 0.009 & -0.010 & 0.006 & 0.021 & 0.000 & 0.004 & 0.029 \\
1510 & 0.677 & 0.774 & 0.707 & 0.620 & 0.684 & 0.744 & 0.762 & 0.020 & 0.295 & 0.000 & 0.347 & \bfseries 0.831 \\
1480 & -0.072 & -0.031 & 0.043 & 0.031 & 0.034 & 0.029 & \bfseries 0.048 & 0.031 & 0.006 & -0.001 & -0.020 & 0.033 \\
11 & 0.139 & 0.135 & 0.140 & 0.147 & 0.130 & 0.140 & 0.120 & 0.076 & 0.097 & 0.000 & 0.083 & \bfseries 0.296 \\
37 & 0.100 & 0.013 & \bfseries 0.211 & 0.205 & 0.181 & 0.179 & 0.121 & 0.110 & 0.050 & 0.000 & 0.077 & 0.200 \\
469 & 0.004 & 0.003 & 0.007 & 0.006 & 0.007 & 0.007 & 0.006 & 0.006 & \bfseries 0.009 & 0.001 & 0.006 & 0.006 \\
458 & 0.951 & 0.959 & 0.695 & 0.766 & 0.650 & 0.705 & 0.747 & \bfseries 0.996 & 0.178 & 0.060 & 0.400 & 0.786 \\
1464 & 0.034 & -0.048 & 0.077 & \bfseries 0.126 & 0.061 & 0.066 & 0.013 & 0.014 & 0.009 & 0.000 & 0.024 & 0.079 \\
1068 & \bfseries 0.184 & 0.060 & 0.157 & 0.176 & 0.105 & 0.110 & 0.025 & 0.030 & 0.080 & 0.000 & 0.014 & 0.148 \\
1049 & 0.090 & 0.080 & 0.049 & 0.064 & 0.055 & 0.056 & 0.098 & 0.058 & 0.051 & 0.008 & 0.011 & \bfseries 0.105 \\
23 & 0.027 & 0.003 & 0.026 & 0.032 & \bfseries 0.035 & 0.031 & 0.027 & -0.007 & 0.018 & 0.000 & -0.003 & 0.033 \\
1050 & 0.046 & \bfseries 0.116 & -0.048 & 0.109 & -0.020 & -0.033 & 0.025 & 0.006 & 0.060 & -0.000 & -0.057 & 0.073 \\
40975 & 0.013 & 0.013 & 0.011 & 0.021 & 0.011 & 0.011 & 0.013 & \bfseries 0.110 & 0.021 & 0.000 & 0.029 & 0.035 \\
40982 & 0.156 & 0.142 & 0.217 & 0.207 & 0.218 & 0.221 & 0.125 & \bfseries 0.223 & 0.082 & 0.007 & 0.192 & 0.163 \\
1067 & 0.209 & 0.162 & 0.185 & 0.185 & \bfseries 0.218 & \bfseries 0.218 & 0.058 & 0.113 & 0.119 & 0.105 & 0.179 & 0.211 \\
\midrule
Avg. Rank & 6.4 (3.1) & 7.4 (3.6) & 5.1 (3.0) & 3.6 (1.9) & 5.2 (2.8) & 5.2 (2.6) & 6.8 (3.3) & 6.4 (3.7) & 7.4 (3.2) & 11.1 (1.3) & 8.8 (2.6) & \bfseries 2.8 (1.7) \\
\midrule
Overall Rank & 6 & 10 & 3 & 2 & 5 & 4 & 8 & 7 & 9 & 12 & 11 & \bfseries 1 \\
\bottomrule
\end{tabular}
}
\end{table}
\begin{table}[h]
\vspace{-10pt}
\caption{Time required in seconds to complete 1000 epochs for clustering dataset 1510. The relative time is the computation time of other baseline methods relative to the time taken by the proposed method (relative time 1.0).}
\label{tab:time_comparison}
\centering
\scalebox{0.70}{
\begin{tabular}{lrrrrrrrrrrrr}
\toprule
Method & K-means(X) & GMM(X) & GMM(Z) & K-means(Z) & DEC & G-CEALS & IDEC & DKM & AECM & DEPICT & DCN & DynAE \\
\midrule
Time (Seconds) & 0.06 & 0.07 & 40.08 & 40.13 & 71.29 & 73.16 & 84.58 & 97.09 & 268.94 & 307.17 & 384.91 & 1718.55 \\
Relative time & 0.00 & 0.00 & 0.55 & 0.55 & 0.97 & 1.00 & 1.16 & 1.33 & 3.68 & 4.20 & 5.26 & 23.49 \\
\bottomrule
\end{tabular}
}
\end{table}

\subsection {Clustering of tabular data sets}
Table~\ref{tab:results_with_acc} presents the clustering accuracy (ACC) and rank ordering of the baseline and proposed methods. Similar to other studies on tabular data, no single method performs the best on all data sets due to data heterogeneity. The AE-CM method yields superior clustering accuracy on data sets with IDs 40994, 1480, 1464, 1068, 1049, and 40975. However, this method results in some of the lowest accuracy scores on other datasets. Similarly, the DynAE method outperforms all methods on three data sets with ID 458, 1050, and 40982. Our proposed G-CEALS method outperforms all baselines on four data sets with ID 1063, 1510, 11, and 37.

Therefore, the tabular data literature commonly uses rank ordering to demonstrate the generalizability of a learning algorithm. Our proposed G-CEALS method shows the best average rank of 2.9 (1.7) across 16 tabular data sets, outperforming all other competitive deep clustering baselines DCN (5.1 (3.7)) and AE-CM (5.6(4.8)).  These results are important because traditional clustering methods have long been used as de facto methods for tabular data. A GMM clustering on Z space (GMM (Z)) is outperformed by only two deep clustering methods (DCN and proposed G-CEALS).

The rank-ordering results based on clustering accuracy are consistent with those obtained using ARI scores. Table~\ref{tab:results_with_ari} ranks our proposed deep clustering method as the best (an average rank of 2.8(1.7)) among all methods based on ARI scores. Although the AE-CM method yields competitive clustering accuracies (ACC) on multiple data sets (Table~\ref{tab:results_with_acc}), its ARI scores are nearly zero in most cases. For almost all other baseline methods, at least one data set yields a negative ARI score, which indicates a discordance between predicted cluster labels and ground truth labels.  In contrast, none of the ARI scores obtained by the proposed G-CEALS method is negative.

\subsection {Time complexity}
One of the reasons traditional clustering methods have been a \emph {de facto} choice for tabular data is computational time. Even at the expense of heavy computations, deep learning methods have not shown great success in outperforming traditional machine learning on tabular data. Table~\ref{tab:time_comparison} presents the training times for all methods when using the dataset with ID 1510. Relative to other competitive deep clustering methods (DCN, AE-CM), the proposed G-CEALS requires three to five times less computational time. However, a little over a minute computational time yields up to 64\% increase in clustering accuracy compared to baseline methods with faster computational time (K-means (X), GMM (X), GMM (Z), K-means (Z), DEC). Therefore, the proposed G-CEALS method provides superior clustering accuracy at a reasonably low computational cost.
\section {Discussion of results}

The paper proposes one of the first deep clustering methods for tabular data when a recent survey on deep clustering methods suggests no work on such data~\citep{zhou2022comprehensive}. The key findings of this article are as follows. First, the proposed G-CEALS method is superior to eleven baseline traditional and deep clustering methods in average ranks across sixteen tabular data sets. Second, the proposed method demonstrates effective cluster separation on deep feature space by learning Gaussian cluster parameters whereas existing models learn the mean of t-distributed clusters. Third, the proposed method handles cluster imbalance problem in tabular data by learning individual cluster weights instead of assuming balanced clusters. Fourth, the proposed method shows faster computational cost compared to other competitive deep clustering methods. Even when the proposed methods is computationally more expensive than some traditional clustering methods, it offers superior clustering accuracy.

\subsection{Traditional versus deep clustering}
A general observation in the deep classification of tabular data studies is traditional machine learning of input features (X) is often superior to deep learning methods~\citep{Kadra2021,Shwartz-Ziv2022}. However, a simple autoencoder learned embedding (Z) achieves better clustering performance than traditional machine learning of X. Among the deep clustering methods, DCN amd AC-EM methods show superior clustering accuracy to traditional clustering methods. However, ARI scores reveal that traditional clustering (K-means or GMM) on autoencoder learned embedding (Z) is superior to all baseline deep or traditional clustering methods (K-means or GMM on X). It is known that an accuracy metric may not be reliable in the presence of data imbalance. Data imbalance problem is usually not seen as an issue with image data sets as it is with tabular data sets. In this context, metrics like ARI may reveal important insights into cluster randomness. Therefore, current deep clustering methods (AE-CM, DKM, DCN), benchmarked on or developed for image data sets, may not yield robust clustering performance on tabular data compared to more traditional methods (GMM(Z), K-means (Z), DEC, IDEC). In contrast, proposed G-CEALS method achieves the best performance on both clustering accuracy and ARI scores, indicating its effectiveness on tabular data.


\subsection {Image versus tabular data embedding}

Traditional clustering methods are obsolete in computer vision because clustering on high-dimensional homogeneous pixel space is ineffective. In contrast, tabular data sets have smaller sample sizes and dimensionality with heterogeneous features, whereas traditional clustering is still relevant and effective. In this context, our results reveal that deep architectures with convolutional neural networks are not as effective in learning tabular data embedding as they are with image data. Our observation confirms a preliminary study showing that state-of-the-art deep clustering methods optimized for image data sets do not produce satisfactory clustering accuracy on tabular data~\citep{Sakib2023effectiveness}. This suggests the need for specialized learning algorithms and architectures for tabular data similar to the proposed G-CEALS method.

\subsection{Effects of data statistics on clustering performance}


Tabular data sets are said to be heterogeneous because of heterogeneity in feature space and data statistics. We discuss three scenarios in this context. First, categorical features extend the data dimension with additional one-hot encoded binary columns. Three (IDs 469, 23, and 40975) of the sixteen tabular data sets have categorical features only. The proposed G-CEALS is the best for categorical tabular data (average rank 2.7 (0.6) followed by other deep learning (average rank 6.5 (3.5)) and traditional clustering (average rank 7.3 (1.8)) methods. For twelve numerical-only tabular data sets, average rank orderings are similar, G-CEALS (3.0 (2.0), other deep learning 6.6 (3.5), and traditional clustering 7.0 (3.4). The AE-CM achieves the best clustering accuracy (71.3) for only one mixed data set (ID 1480) but suffers a negative ARI score (-0.001). Conversely, G-CEALS shows a better balance in clustering accuracy (65.2) and ARI (0.033).

Second, a higher F-S ratio indicates wider data tables, whereas a lower F-S ratio refers to taller data tables. Tabular data sets with a higher F-S ratio are more likely to introduce the \emph {curse of dimensionality} on machine learning. For datasets with an F-S ratio below 1.0, G-CEALS achieves an average rank of 2.0 (1.0), whereas other deep and traditional methods rank 6.4 (3.6) and 8.2 (2.8), respectively. Conversely, the average ranks for datasets with an F-S ratio above 1.0 are 3.2 (1.8) for G-CEALS, 6.6 (3.5) for other deep learning methods, and 7.1 (3.1) for traditional clustering.

Third, the C-score measures between-feature correlations of a tabular data set, which may affect machine learning performance. Tabular data sets with high C-scores ($>$0.10) are best clustered using the G-CEAL method (average rank 3.3 (1.9)), whereas the G-CEAL method ranks (2.3 (1.2)) the best on tabular data sets with low C-scores ($<$ 0.10). The average rank orderings of deep learning methods are 6.6(3.3) and 65. (3.6) on low and high C-score data sets, respectively. Traditional clustering methods show inferior rank orderings of 7.5(3.1) on low and 7.2(3.6) on high C-score data sets.

\subsection{Ablation study}

\begin{figure*}[t]
\vspace{-10pt}
\subfigure[Accuracy] { \includegraphics[trim=0 0.5cm 0 0cm, clip, width=0.33\textwidth] {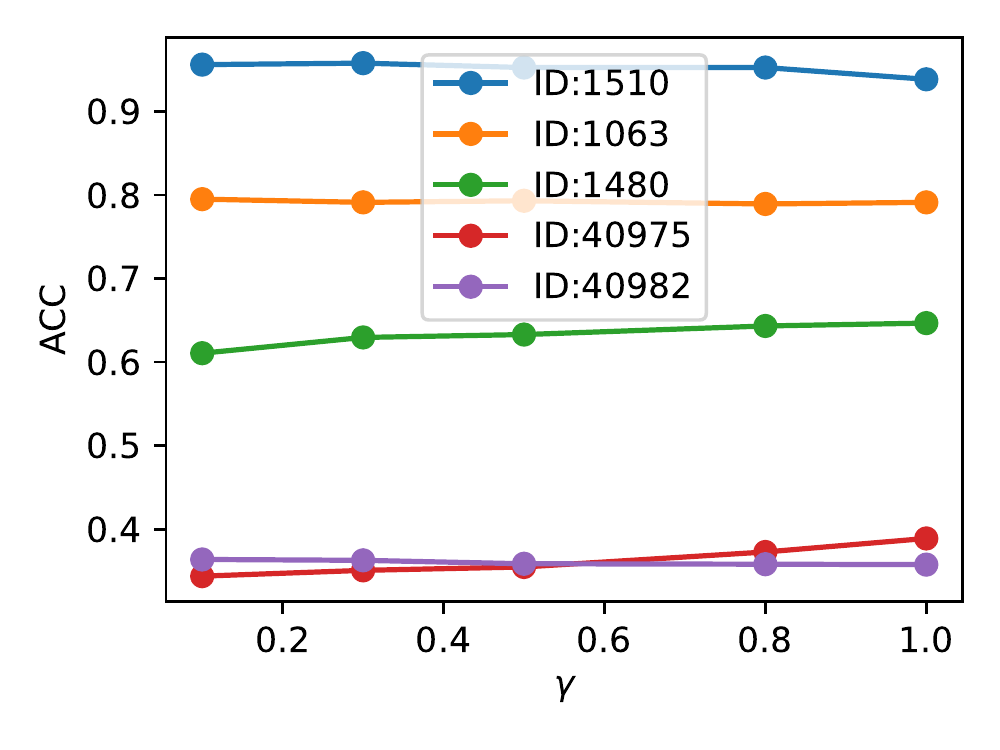}}
\hspace{-10pt}
\subfigure[ARI] { \includegraphics[trim=0 0.5cm 0 0cm, clip, width=0.33\textwidth] {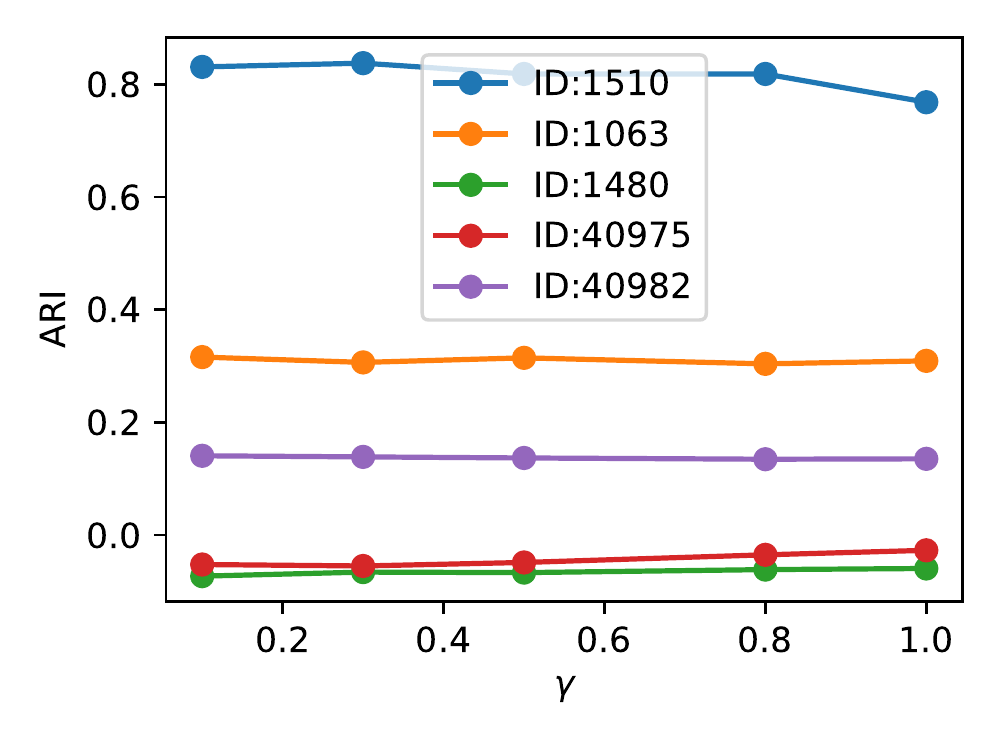}}
\caption{Effects of $\gamma$ values on the clustering accuracy and ARI scores of the proposed G-CEALS methods using five tabular data sets.
}
\vspace{-10pt}

\label{fig:ablation_gamma}
\end{figure*}

The effect of the clustering loss balancing parameter $\gamma$ on clustering accuracy and ARI scores is shown in Figure~\ref{fig:ablation_gamma}. The results indicate that clustering performance remains relatively stable across varying $\gamma$ values. However, the choice of $\gamma$ value can impact the time and stability of convergence, although a delayed convergence is expected to yield similar clustering performance. Therefore, a $\gamma$ value in the lower range is preferred to ensure stable convergence and clustering performance.

\subsection {Limitations}

Despite promising clustering performance, the proposed G-CEALS method has several limitations as any other method. Unsupervised learning or clustering is not trivial via deep learning, which often expects a target variable or a robust learning objective. There is still plenty of room to improve the cluster performance by innovating novel learning objectives or targets. Furthermore, data or cluster imbalance is common with tabular data sets, which would require more algorithmic solutions instead of the proposed early stopping to avoid the clusters from merging. Several aspects of model selection (e.g., embedding dimensions and network architecture) vary due to the heterogeneity of tabular data sets, unlike image data sets in computer vision applications. The proposed method needs better approaches to model selection and optimization.


\section {Conclusions}

This paper presents a novel deep clustering method for concurrently learning cluster-friendly embedding and cluster assignments of unlabeled tabular data. The superiority of the proposed G-CEALS method against nine number of state-of-the-art clustering methods suggests that multivariate Gaussian distributions learn clusters better than the widely used t-distribution. Furthermore, dynamically updating the target cluster distribution is more effective than setting a closed-form target for deep clustering. The cluster weight is important in ensuring proper cluster separation during cluster imbalance. A data-informed decision is recommended when selecting an appropriate clustering approach because one method may not be suitable for all tabular data. The proposed deep clustering shows a promising approach that may replace traditional machine learning approaches for clustering tabular data.

\section*{Acknowledgements}
The research reported in this publication was supported by the Air Force Office of Scientific Research under Grant Number W911NF-23-1-0170. The content is solely the responsibility of the authors and should not be interpreted as representing the official policies, either expressed or implied, of the Army Research Office or the U.S. Government.

\bibliographystyle{unsrt}
\bibliography{mybib}

\end{document}